\documentclass[12pt]{article}

\usepackage{arxiv}
\usepackage[utf8]{inputenc} %
\usepackage[T1]{fontenc}    %
\usepackage{hyperref}       %
\usepackage{url}            %
\usepackage{booktabs}       %
\usepackage{amsfonts}       %
\usepackage{nicefrac}       %
\usepackage{microtype}      %
\usepackage{lipsum}
\usepackage{natbib}
\usepackage{xcolor}         %
\usepackage{multirow}
\usepackage{makecell}
\usepackage{comment}
\usepackage{wrapfig}
\usepackage{amsmath}
\usepackage[inline]{enumitem}
\usepackage{arydshln}
\usepackage[most]{tcolorbox}
\usepackage{subcaption}
\usepackage{graphicx}
\usepackage{hyperref} %
\usepackage{float}
\usepackage{lipsum}
\usepackage{adjustbox}
\usepackage{framed}
\usepackage{xspace}
\usepackage{comment}
\usepackage{bbm}
\usepackage{diagbox}
\usepackage{authblk}

\newcommand{\model}{\texttt{Be.FM}\xspace}

\title{\model: Open Foundation Models for Human Behavior}

\makeatletter
\renewcommand\AB@affilsepx{\quad\protect\Affilfont}
\makeatother
\renewcommand*{\Affilfont}{\normalsize\normalfont}

\author[.]{
    Yutong Xie$^{1*}$ \,
    Zhuoheng Li$^{1*}$ \,
    Xiyuan Wang$^{1*}$ \,
    Yijun Pan$^1$ \,
    Qijia Liu$^1$ \,
    Xingzhi Cui$^1$ \,
    Kuang-Yu Lo$^1$ \, \\
    Ruoyi Gao$^1$ \,
    Xingjian Zhang$^1$ \,
    Jin Huang$^1$ \,
    Walter Yuan$^2$ \,
    Matthew O. Jackson$^{3,4}$ \,
    Qiaozhu Mei$^1$
}

\affil[1]{University of Michigan}
\affil[2]{MobLab}
\affil[3]{Stanford University}
\affil[4]{Santa Fe Institute}
\makeatletter
\renewcommand\AB@affilsepx{\\\protect\Affilfont}
\makeatother

\affil[*]{\textit{Equal contribution}}
\affil[1]{\texttt {\{yutxie,zhlii,denniswx,panyijun,ponypony,tigercui,lonick\}@umich.edu}}
\affil[1]{\texttt {\{gaoruoyi,jimmyzxj,huangjin,qmei\}@umich.edu}}
\affil[2]{\texttt{\{walter.yuan\}@moblab.com}}
\affil[3]{\texttt{\{jacksonm\}@stanford.edu}\vspace{10pt}}

\begin{document}
\maketitle

\begin{abstract}
Despite their success in numerous fields, the potential of foundation models for modeling and understanding human behavior remains largely unexplored. We introduce Be.FM, one of the first open foundation models designed for human behavior modeling. Built upon open-source large language models and fine-tuned on a diverse range of behavioral data, \model can be used to understand and predict human decision-making. %
We construct a comprehensive set of benchmark tasks for testing the capabilities of behavioral foundation models. Our results demonstrate that \model can predict behaviors, infer characteristics of individuals and populations, generate insights about contexts, and apply behavioral science knowledge. %
The models are available upon request: \href{https://forms.gle/DAvxJYReqg7midQn9}{\texttt{LINK}}. 
\end{abstract}

\section{Introduction}

Foundation models have revolutionized numerous fields. Trained on vast and diverse datasets, these models exhibit remarkable adaptability across a wide range of simple and complex tasks. Applications built on these models now support both everyday and specialized tasks with performance approaching human experts. For instance, large language models (LLMs) assist in writing, communicating, and information seeking; multi-modal models enable rapid generation of images, videos, and music; and scientific foundation models provide literature reviews, code generation, data analysis, and pattern recognition \citep{bubeck2023sparks}.  In fields such as genomics, foundation models like AlphaFold have been used to map transcriptional activity across human cell types, providing new insights into gene regulation and advancing our understanding of biological processes \citep{jumper2021highly}. Integrating general language understanding with domain-specific reasoning abilities, foundation models have evolved beyond mere analytical tools. They now serve as catalysts for innovation, driving new discoveries and interdisciplinary collaborations.

Despite their transformative potential, foundation models remain largely underexplored for modeling \textit{human behavior} and addressing problems in \emph{behavioral science}---the analysis and prediction of human behaviors. Behavioral science spans multiple disciplines, including psychology, sociology, anthropology, and economics, and serves as a foundation for understanding decision-making dynamics at individual, group, and societal levels. Insights from this field inform policymaking, healthcare, education, and business strategies, highlighting the need for more advanced AI-driven approaches to enhance behavioral analysis and intervention.  

Traditional AI models used in behavioral science are often trained on task-specific datasets and handcrafted features \citep{turgeon2020tutorial,kliegr2020advances,dehghan2022applications}, limiting their ability to generalize across diverse contexts and tasks. Machine learning models built for specific tasks like sentiment analysis \citep{braig2023machine}, survey analysis \citep{heeringa2017applied}, and clickthrough prediction \citep{yang2022click} typically require extensive domain-specific tuning and struggle to capture the full complexity of human behavior.

A behavioral foundation model has the potential to overcome these limitations by leveraging vast and diverse datasets to develop a more comprehensive and generalizable understanding of human behavior. This allows the model to perform effectively across a wide range of tasks without requiring specialized training for each different application. Such a model could enhance the representation and application of behavioral knowledge, improve behavioral prediction and outcome estimation, provide deeper insights into how context shapes behavior, uncover latent motivations, and enable scalable analysis at both individual and population levels.

Developing a foundation model for human behavior presents two key challenges. 
First, the fragmented and dispersed nature of behavioral data makes it difficult to organize a diverse and high-quality portfolio of datasets for effective model training. 
Second, the lack of standardized benchmark tasks hinders the systematic evaluation and refinement of model performance, limiting its reliability and applicability across different behavioral science domains and tasks.

\begin{figure}[t]
    \centering
    \includegraphics[width=0.9\linewidth]{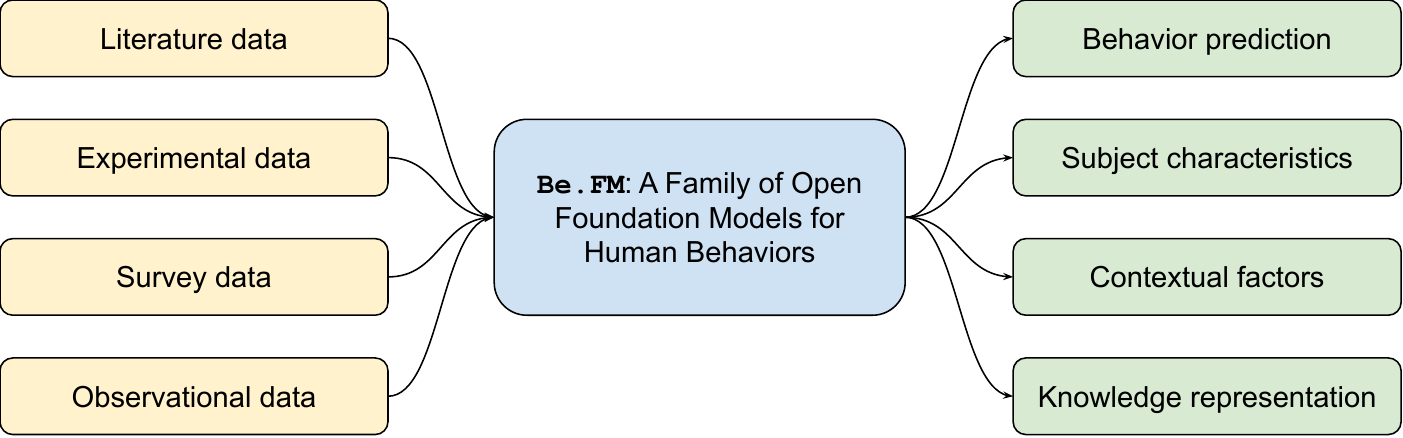}
    \caption{\model is a foundation model designed for modeling human behavior. 
    Trained on a diverse portfolio of behavioral datasets, \model has the capabilities to:
    (1) Predict and simulate behavior across diverse scenarios;
    (2) Infer motivations and other characteristics of individual humans as well as broader populations from their behaviors;
    (3) Generate insights about contextual factors that influence human behaviors; and 
    (4) Represent and apply behavioral knowledge in reasoning and problem-solving. 
    }
    \label{fig:fm-illus}
\end{figure}

We introduce \model, a family of foundation models specifically designed for modeling and emulating human behavior (Figure \ref{fig:fm-illus}). Built upon the powerful language capability of open-source large language models, \model models are trained on curated datasets encompassing multiple key categories of behavioral data: behavioral science literature, human-subject experimental data, survey data, and observational data (Sec. \ref{sec:model}). 
We introduce a comprehensive suite of benchmark tasks,  aligned with critical needs of behavioral science, to systematically evaluate the effectiveness of behavioral foundation models (Sec. \ref{sec:tasks}). 
Through extensive experiments, we demonstrate that \model has obtained a broad range of capabilities to:
(1) \textbf{Predict and simulate behavior across diverse scenarios} (Sec. \ref{sec:infer-behavior}); 
(2) \textbf{Infer characteristics of individual subjects and populations} (Sec. \ref{sec:infer-subject});
(3) \textbf{Generate insights about contextual factors that influence behavior} (Sec. \ref{sec:infer-context}); and 
(4) \textbf{Represent behavioral science knowledge and apply it. %
} (Sec. \ref{sec:infer-knowledge}). 
Through these capabilities, \model provides a foundation for advancing behavioral science by enabling more comprehensive and scalable analysis. Additionally, it introduces a novel interface that integrates AI with human behavior research and applications, %
as well as real-world deployment. %

\section{Overview of \model}
\label{sec:model}

\model is %
built upon the open-source large language models (e.g., Meta Llama 3.1). \model models are further trained on diverse sources of behavioral data to enhance their ability to address a wide range of human behavior-related tasks. The foundation of \model is that human behavior is shaped by multiple factors, including the  environment and context of the behavior, the characteristics of the subject, and underlying motivations. The framework can be formally represented as
\begin{equation}
y = \mathcal{F}(\mathbb{K}, x, c),
\label{eqn:framework}
\end{equation}
where $x$ represents the set of characteristics of the human subject,
$c$ denotes the set of contextual variables that defines the environment,
$\mathbb{K}$ encompasses all relevant knowledge about human behaviors, and
$\mathcal{F}$ is a latent function that maps these inputs to a specific behavioral choice 
$y$. The training and benchmarking of \model are guided by this framework.

The rest of this section outlines the behavioral data used for training (Sec. \ref{sec:data}) and the training process (Sec. \ref{sec:train}). 
Additional details, including the processing and formatting of the training data, are provided in Appendix \ref{app:train}.

\subsection{Behavioral Data}
\label{sec:data}

\begin{table}[t]
    \centering
    \def\arraystretch{2.0}
    \begin{tabular}{p{3cm}p{6cm}p{6cm}}
    \toprule
    \textbf{Data Category} & \textbf{Purpose} & \textbf{Example Data Description} \\
    \midrule
    \textbf{Literature Data} & To provide foundational knowledge and reasoning clues about human behavior. Literature data provide theories, empirical findings, established mechanisms, and the evolution and trends of behavioral science research. 
    A proxy of $\mathbb{K}$ as in Equation~\ref{eqn:framework}. & Publications from behavioral science related journals. %
    So far, titles and abstracts from 2,703 publications have been collected and augmented with extracted research workflows, resulting a corpus of 3,129,462 tokens.\\ %
    \textbf{Human-Subject Experimental Data} & To document behavior of human subjects in controlled environments, providing empirical observations on human decision-making, social interactions, and behavioral responses to specific conditions. A proxy of $\{x, c, y\}$ as in Equation~\ref{eqn:framework}. & Experiment log from human-subject experiments. This version includes data from MobLab, containing game-playing records from 68,780 subjects, 82,057 independent observations, and 15,236 sessions in multiple economic games, collected over a nine-year period (2015–2023).\\
   \textbf{Survey Data} & To capture  personality traits, attitudes, behavioral tendencies, and other self-reported information related to personal behavior. Surveys can inquire about $x$ or $y$ in Equation~\ref{eqn:framework}, often under pre-framed context $c$ in survey questions. & Responses to surveys and summarized results, such as the Big Five personality test, which covers a population of 17,667 subjects and a total of 883,350 responses to survey questions. \\
    \bottomrule
    \end{tabular}
    \vspace{5pt}
    \caption{A variety of behavioral data used in training \model. }
    \label{tab:data}
    \vspace{-10pt}
\end{table}

Training a behavioral foundation model requires a wide spectrum of high-quality behavioral data. However, existing data are often fragmented, scattered across disciplines, and not structured for integration into a language model. To address this challenge, we identify four key categories of behavioral data (Table \ref{tab:data}) that together provide a high coverage of variables in Equation~\ref{eqn:framework} and discuss how they can contribute to the overall capabilities of \model.

\paragraph{Literature data. }
The behavioral science literature documents existing theories and scientific findings about human behavior, forming a proxy of the behavioral knowledge $\mathbb{K}$ in Equation~\ref{eqn:framework}. To train the initial version of \model, we use publications from \textit{American Economic Review}. %
Our dataset includes a curated selection of 3,003 publications from this journal, ranging from the year 1963 to 2024. Among these, 2,703 publications are used for training the foundation model, while the remaining ten percent are held out for model evaluation. To precisely extract and represent the behavioral knowledge, we first summarize the title and abstracts of each study into a structured research workflow: context of research, key idea, method, outcome, and projected impact, following the practice in \citet{zhang2024massw}, resulting in a total of 3,129,462 tokens. %
Trained to predict the title of a paper using the research workflow and to predict the key idea given the context of research, \model incorporates key concepts, theories, methodologies, and empirical findings in behavioral sciences, strengthening its ability to complete tasks such as behavior prediction, context inference, research planning, and solving problems like the Economics Olympiad questions. %

\paragraph{Experimental data}
Experimental data captures behavioral patterns of diverse subject pools in controlled environments, offering key insights into individual and group decision-making, as well as social interactions. Such data is usually represented in the format of $\{x, c, y\}$, where $x$ represents subject characteristics (often partially labeled),
$c$ denotes contextual variables (usually partially identified, often including the experiment condition), and 
$y$ records the observed behavior. Valuable sources of such data are experimental platforms and published repositories that record the responses of subjects in various behavioral, economic, and psychological studies.
In the initial version, we utilize a collection of experiments from  MobLab (\cite{mei2024turing}), a platform for behavioral science and economic games, which includes game-playing experiment data from 68,779 subjects and 82,057 independent observations collected over a nine-year period (2015–2023). Individual sessions in this data are used for model training, and a different task is used for evaluation, to predict the behavioral distributions at the population level. 
Trained to predict individual behaviors given game conditions, \model captures behavioral tendencies across diverse scenarios, enhancing its accuracy in behavior prediction and alignment, which enables improved simulation and analysis of real-world behavioral patterns.

\paragraph{Survey data}
Survey data provides self-reported insights into personality traits, attitudes, and preferences. They mainly offer a representation of the characteristics of subjects ($x$ in Equation~\ref{eqn:framework}), although they sometimes also include self-reported behaviors $\Tilde{y}$ which provide a proxy of the observed behaviors $y$, and can also be useful in simulating responses to surveys that have not been run yet. These responses may be solicited directly without specifying the context, or with certain pre-specified scenarios $c$ in the survey questions. In the initial version of \model, we use a dataset of responses from the Big Five personality test. %
The dataset includes demographic information (race, age, and gender) of respondents and their single-choice answers to 50 questions across five personality dimensions: openness, conscientiousness, extraversion, agreeableness, and neuroticism (\cite{mei2024turing}). These responses are aggregated into five dimensional scores of individual personality traits. In total, records from 17,667 subjects are used for training, while 1,963 are reserved for evaluation.
The large survey data enables \model to infer individual behavioral patterns, predict correlations in responses across questions, and emulate subjective experiences. Trained to predict the personality traits of subjects given their demographic information, these data enhance \model's ability to generalize across demographic groups, improving behavioral modeling at both individual and population levels.

\paragraph{Observational data}

Observational data captures real-world behaviors by capturing unstructured or semi-structured actions in natural or semi-natural settings. It therefore overlaps with experimental data and survey data and provides a similar representation $\{x, c, y\}$, although usually associated with a much wider and noisier representation of $x$ and $c$. Examples include activity logs from online platforms and social media, search and browsing patterns, and video recordings of group interactions. These datasets are rich in variability, reflecting spontaneous, less controlled behaviors, which are valuable for generalizing to complex, real-life scenarios. Using observational data to train the behavioral foundation model follows a similar procedure as experimental data, promoting additional adaptability and robustness. %
Observational data have not used to train the initial version of \model, %
although they will be included in future versions. 

\subsection{Model Configuration and Training}
\label{sec:train}

To train the initial versions of \model, we adopt two releases of Meta Llama 3.1 (70B and 8B) \citep{grattafiori2024llama} as the foundation LLMs. To improve computational efficiency, the 70B model is quantized with 8-bit precision using the bitsandbytes library, reducing memory overhead while preserving performance.
We fine-tune the model using LlamaFactory \citep{zheng2024llamafactory}, employing supervised fine-tuning (SFT) with low-rank adaptation (LoRA) \citep{hu2022lora} to enable efficient weight updates. LoRA is applied across all model layers, ensuring fine-tuning without excessive computational costs.
Training data is formatted following the Alpaca template \citep{taori2023alpaca}, aligning inputs with the model’s expected structure. Preprocessing includes tokenization, truncation, and caching, with multiprocessing used to improve scalability and reduce latency. This fine-tuning approach optimizes \model for behavioral science applications while maintaining efficiency and flexibility.

\begin{table}[t]
    \centering
    \def\arraystretch{1.5}
    \begin{tabular}{|p{3.5cm}|p{7cm}|p{4cm}|}
    \hline
    \textbf{Capability} & \textbf{Tasks} & \textbf{Evaluation Data} \\
    \hline
    \multirow{2}{*}{\parbox{3.5cm}{\vspace{8pt}\textbf{Predict and simulate \\behavior across\\ diverse scenarios}}}
    & Predicting behavior in experiments, 
    \newline given one/multiple of the following:
    \begin{itemize}[leftmargin=12pt]
        \vspace{-4pt}
        \setlength{\itemsep}{-2pt}
        \item Subject information such as demographics
        \item Contexts such as experiment design
        \vspace{-10pt}
    \end{itemize} & Human-playing records in multiple behavioral economics games on MobLab \\ \cline{2-3} %
    & Predicting responses to survey questions, 
    \newline given one/multiple of the following:
    \begin{itemize}[leftmargin=12pt]
        \vspace{-4pt}
        \setlength{\itemsep}{-2pt}
        \item Subject information such as demographics
        \item Contexts such as survey design
        \vspace{-10pt}
    \end{itemize} & BigFive test responses \\ \hline
    \multirow{2}{*}{\parbox{3.5cm}{\vspace{8pt}\textbf{Infer characteristics \\about subjects}}}
    & Predicting subject demographics, 
    \newline given one/multiple of the following:
    \begin{itemize}[leftmargin=12pt]
        \vspace{-4pt}
        \setlength{\itemsep}{-2pt}
        \item Behavior in experiments
        \item Survey responses
        \vspace{-10pt}
    \end{itemize} & BigFive test responses \\ \hline %
    \multirow{2}{*}{\parbox{3.5cm}{\textbf{Infer contextual \\factors}}}
    & Predicting experiment interventions, 
    \newline given one/multiple of the following:
    \begin{itemize}[leftmargin=12pt]
        \vspace{-4pt}
        \setlength{\itemsep}{-2pt}
        \item Observed changes in behaviors
        \item Subject/population information
        \vspace{-10pt}
    \end{itemize} & Dictator game meta-analysis \\ \hline %
    \multirow{2}{*}{\parbox{3.5cm}{\textbf{Apply knowledge in \\ reasoning and problem solving}}}
    & Research workflow reasoning & Publications from \emph{American Economic Review} \\ \cline{2-3}
    & Complex Problem-solving & IEO contest problems \\ \hline

    \end{tabular}
    \vspace{5pt}
    \caption{General capability of \model in various behavioral science tasks. 
    }
    \label{tab:benchmark}
\end{table}

\section{Generalizable Capability of \model in Various Behavioral Science Tasks}
\label{sec:tasks}

Behavioral science encompasses a wide range of tasks, such as survey analysis, behavioral prediction, and simulations of interactions, which can be applied across populations and contexts. 
Traditional tools and AI models in this field often rely on task-specific datasets and handcrafted features \citep{turgeon2020tutorial,kliegr2020advances,dehghan2022applications}, limiting their ability to generalize across different applications. 
Trained with a wide spectrum of behavioral data, a behavioral foundation model has the potential to overcome these constraints and perform more effectively across various applications, and do better in novel scenarios. Also, it may hold a broader understanding of human behavior especially at the population level. 
We employ a comprehensive series of benchmark tasks designed to capture key dimensions of behavioral science, enabling a systematic evaluation of \model’s capabilities. These tasks and corresponding capabilities are organized surrounding the general framework in Equation~\ref{eqn:framework} and aligned with the key dimensions of behavioral science, as outlined in Table \ref{tab:benchmark}. 
Additional details--including the test data, prompts, evaluation metrics, and baselines--are provided in Appendix \ref{app:benchmarks}. 

\subsection{\model's ability to predict and simulate behaviors across scenarios}
\label{sec:infer-behavior}

Traditional methods for behavior prediction often struggle to generalize across diverse populations and settings \citep{muthukrishna2019problem,yarkoni2022generalizability,d2022underspecification,barocas2023fairness}. In contrast, the \model foundation model leverages large-scale data and contextual reasoning, enabling more adaptive and accurate inferences across real-world scenarios.

\paragraph{Predicting behaviors as a function of characteristics and settings in economic games. }
A key aspect of behavior prediction is the ability to infer behaviors or experiment outcomes at the population level, tracing out a distribution as a function of characteristics and settings. %
Building on prior work \citep{mei2024turing, xie2024different}, we evaluate \model using five classic economic games, covering seven distinct scenarios: (1) the Dictator Game, (2) the Ultimatum Game (as both Proposer and Responder), (3) the Trust Game (as both Investor and Banker), (4) the Public Goods Game, and (5) the Bomb Game. These games assess various behavioral tendencies, including altruism, fairness, trust, and risk preference. 
Formally, the prediction task can be represented as 
\begin{equation}
    X, c \rightarrow \mathbf{E}_{x \in X}[p(y|x, c)],
\end{equation}
where $c$ is a given game scenario, $X$ is a population of human subjects, and $\mathbf{E}[p(y|x, c)]$ is the expected behavioral choice as a function of this subject population characteristics in this game scenario. 

\begin{figure}[t]
    \centering
    \includegraphics[width=\linewidth]{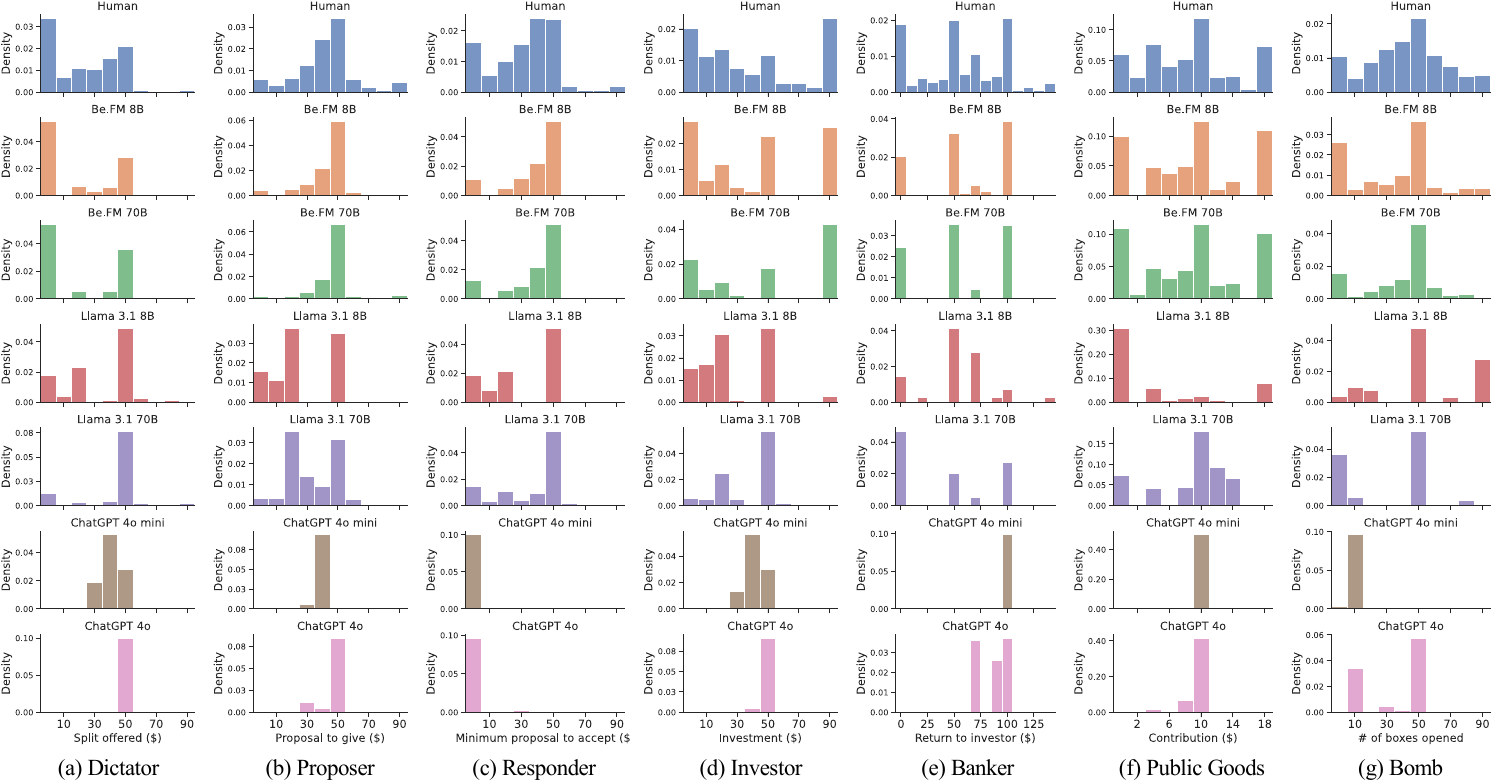}
    \caption{
    Behavior distributions of human players (the first row) and models (remaining rows) in classic behavioral economic games. 
    Each column of histograms represents a specific game scenario. 
    \model (the second and third rows) narrows the gap between AI-generated and human behavior distributions, demonstrating improved alignment and more accurate population-level behavior simulation. 
    }
    \label{fig:histograms-moblab}
\end{figure}

For each game, we generate 1,000 independent samples by using the game instructions as prompts, and compare the resulting distributions to human actions recorded on the MobLab platform. 
Figure \ref{fig:histograms-moblab} presents behavior histograms comparing human players and evaluated models. The results reveal a significant misalignment between the actual human behavior distribution and the predictions of the base LLMs (Llama 3.1) and other general-purpose LLMs, including a frontier commercial model GPT 4o. However, trained through extensive behavioral data, \model reduces this gap and more accurately simulates human behavior distributions.

\begin{table}[t]
\centering
\begin{tabular}{lccccccc} 
\toprule
\textbf{Model} & \textbf{Dictator} & \textbf{Proposer} & \textbf{Responder} & \textbf{Investor} & \textbf{Banker} & \textbf{Public Goods} & \textbf{Bomb} \\
\midrule
\model 8B   & 6.76  & \textbf{6.94}  & 7.24  & \textbf{4.23} & 10.54  & 3.44  & 9.65  \\ 
\model 70B & \textbf{6.14}  & 7.59  & 6.55  & 13.82 & \textbf{4.31}  & \textbf{3.38}  & \textbf{6.85}  \\ 
\midrule
Llama 3.1 8B  & 8.40   & 16.79 & 5.95  & 15.01 & 6.90   & 24.50  & 11.93 \\
Llama 3.1 70B & 25.94 & 11.83 & \textbf{5.04}  & 18.61 & 11.30  & 11.34 & 14.76 \\ 
GPT 4o-mini & 16.59 & 13.99 & 33.40 & 26.78 & 29.50 & 23.40 & 36.65 \\ 
GPT 4o    & 25.48 & 10.89 & 32.43 & 31.01 & 21.88 & 21.25 & 12.38 \\ 
\bottomrule
\end{tabular}
\vspace{5pt}
\caption{
Performances on simulating human behavior distributions across economic game scenarios. Wasserstein distances between the model-generated distributions and the ground-truth human distributions are measured. A lower distance stands for higher distributional similarity. 
\textbf{Bold} texts highlight the best performance in each game. 
}
\label{tab:results-behavior-moblab}
\end{table}

Table \ref{tab:results-behavior-moblab} reports the Wasserstein distances between model-generated distributions and ground-truth human distributions, where smaller values indicate greater similarity. Behavior distributions predicted by \model are consistently closer to human distributions than the base LLMs (with the exception being the Responder role in a trust game) as well as the commercial LLMs. 
These results highlight \model’s advanced ability to predict and simulate human decisions across diverse scenarios at the population level. %

\paragraph{Predicting Big Five personality traits. }

\model also demonstrates the ability to predict survey outcomes at the individual level. %
We evaluate this using responses from 1,963 subjects who completed the Big Five personality test, held-out from model training. The task assesses how well \model can predict the five personality dimension scores based on demographic information such as race, age, and gender. 
Formally, this task can be represented as 
\begin{equation}
    x, \mathbf{c} \rightarrow g(\mathbf{y}),
\end{equation}
where $x$ is the demographical information of an individual, $\mathbf{c}$ is a set of survey questions, and $g(\mathbf{y})$ is a pre-defined aggregation function of the responses of $x$ to questions in $\mathbf{c}$.  Note that the survey outcomes for a subject, $g(\mathbf{y})$, often provide additional characteristics about the subject and can become part of $x$ in different tasks (e.g., Sec. ~\ref{sec:infer-subject}).

\begin{table}[t]
\centering
\begin{tabular}{lccccc} 
\toprule
\textbf{Model} & \textbf{MAE}($\downarrow$)  & \textbf{Spearman's corr}($\uparrow$) & \textbf{W-distance}($\downarrow$) \\ 
\midrule
\model 8B                      & 7.27 & 0.101$^\ddagger$    & 1.75        \\
\model 70B                     & 7.27 & \textbf{0.128}$^\ddagger$    & \textbf{1.68}$^\dagger$        \\
\midrule
Llama 3.1 8B                   & 8.23 & 0.037   & 5.26        \\
Llama 3.1 70B                  & 7.05 & 0.029   & 4.06        \\
GPT 4o-mini                    & \textbf{6.85} & 0.034   & 5.52        \\
GPT 4o                         & 7.11 & 0.029   & 5.00        \\
\bottomrule
\end{tabular}
\vspace{5pt}
\caption{
Performances on predicting Big Five personality scores given demographics. For each model, the mean absolute error (MAE), Spearman's correlation coefficient, and Wasserstein distance are computed by comparing the predictions and ground truths. Metrics are aggregated over five personality dimensions (OCEAN). 
\textbf{Bold} texts highlight the best performance in each game. 
$^\ddagger$ indicates that the correlation is significant ($p<0.05$). 
``$^\dagger$'' indicates the distributions that pass the smoothed Kolmogorov–Smirnov test ($p>0.05$) with a bin width of 10 for all five dimensions. 
}
\label{tab:results-behavior-bigfive}
\end{table}

Since this task involves individual-level predictions, we adopt two individual-level evaluation metrics in addition to the distribution-level Wasserstein distance: (1) mean absolute error (MAE) and (2) Spearman’s rank correlation. Specifically, for each subject and each personality dimension, the model predicts a score ranging from 10 to 50. MAE is calculated by averaging the absolute errors between the predicted and ground-truth scores. Similarly, Spearman’s rank correlation is computed by correlating the predicted and ground-truth scores for each dimension.

Table \ref{tab:results-behavior-bigfive} presents the results of personality traits prediction, with all results averaged across the five personality dimensions (OCEAN). 
\model obtained  consistently higher correlation and W-distance compared to base models (Llama 3.1) and commercial LLMs. The results on MAE are mixed, indicating that \model has a better ability to preserve the relative orders and the distributions of individual-level predictions rather than the numerical scores. It is worth noting that the mainstream commercial model GPT 4o, although hundreds of times larger in number of parameters, fail to outperform smaller models in predicting individual personality traits. %

\subsection{\model's inference of characteristics of subjects and populations}
\label{sec:infer-subject}

Characterizing individual and groups of subjects is one goal of behavioral science, with applications ranging from understanding behavioral heterogeneity, market segmentation, user profiling, personalized interventions, and improving policy-making. %
To evaluate this capability, we adopt the Big Five personality test data for evaluation and test the model's ability to predict the age of a subject based on their personality traits scores.
Formally, the task can be represented as 
\begin{equation}
    y, (x'), c \rightarrow x, 
\end{equation}
where $x'$ is some partial (often null) information about a subject and $y$ is their behavior under context $c$, and $x$ represents the latent characteristics of the subject that have jointly shaped their behavior $y$.

\begin{table}[t]
\centering
\begin{tabular}{lcccc}
\toprule
\textbf{Model} & \textbf{MAE}($\downarrow$)  & \textbf{Spearman's corr}($\uparrow$) & \textbf{W-distance}($\downarrow$) \\ 
\midrule
\model 8B                      & \textbf{9.05}  & 0.108$^{\ddagger}$     & \textbf{5.39}       \\
\model 70B                     & 9.16  & \textbf{0.134}$^{\ddagger}$     & 5.70       \\
\midrule
Llama 3.1 8B                   & 10.50 & 0.038    & 8.62       \\
Llama 3.1 70B                  & 10.36 & 0.029    & 7.75       \\
GPT 4o-mini                    & 10.17 & 0.030    & 8.07       \\
GPT 4o                         & 11.01 & 0.048$^{\ddagger}$    & 9.12       \\
\bottomrule

\end{tabular}
\vspace{5pt}
\caption{
Performance of models on inferring subjects' ages given Big Five personality scores. 
For each model, the mean absolute error (MAE), Spearman's correlation coefficient, and Wasserstein distance are computed by comparing the model-predicted subject age and the subject reported age. 
\textbf{Bold} texts highlight the best performances.
$^\ddagger$ indicates that the correlation is significant ($p<0.05$).}
\vspace{-5pt}
\label{tab:results-subjects-bigfive}
\end{table}

We use MAE and Spearman’s correlation as the metrics for individual-level evaluation and  Wasserstein distance as the distribution-level metric. 
Table \ref{tab:results-subjects-bigfive} presents the results. %
\model achieves the best results across MAE, Spearman’s correlation, and Wasserstein distance compared with the base models and the commercial LLMs, once again demonstrating that training on behavioral data strengthens its ability to infer things about subjects. The larger non-behavioral model, GPT 4o, again fails to outperform the smaller base models in this task.

\subsection{\model can generate insights about contextual factors}
\label{sec:infer-context}

Behavior is shaped by social, cultural, and environmental contexts, making it essential to incorporate the underlying conditions in which behaviors occur. Inferring the role of context is challenging due to the complexity of external influences. 
A behavioral foundation model may address this challenge by analyzing observed behavioral changes across contexts to infer the most relevant contextual factors. For example, given a shift in behavior, \model can predict situational conditions or experimental settings that may have led to the change. This task can be represented as
\begin{equation}
    X, g(Y), \mathbb{K} \rightarrow c,
\end{equation}
where $X$ is a subject population, $g(Y)$ is a certain utility function defined given the behaviors of the population $Y$ (e.g., a shift of $Y$ compared to the default behavior, or a desired treatment effect), $\mathbb{K}$ is the established knowledge about the behaviors, and $c$ is a contextual variable (e.g., an intervention or a particular experiment design).

To assess \model’s capability in inferring contextual factors, we refer to a meta study of the Dictator Game \citep{engel2011dictator}, which reports various experimental designs alongside their corresponding results. 
We ask \model to suggest treatments that would increase sharing behavior in the Dictator game, and \model outputs experimental designs aligned with those reported in the literature. The recommendations \model made include: (1) introducing social factors, such as providing social norms or expectations \citep{fong2007evidence}, using social identity cues \citep{rigdon2009minimal}, or enhancing social connection \citep{leider2010we}; (2) applying real-world framing \citep{engel2011dictator}; (3) varying cultural and demographic contexts \citep{henrich2005economic}; and (4) incorporating visual cues \citep{rigdon2009minimal}.
Similarly, when predicting treatments that reduce sharing, \model suggests strategies validated by prior studies, including (1) reducing accountability \citep{dana2006you} and (2) altering recipient framing \citep{eckel2005experimental}. Notably, \model also proposes additional treatments not explicitly covered in the meta-analysis, such as introducing time pressure and cognitive load, demonstrating its potential to generate plausible experimental hypotheses beyond the observed data. 
The complete model outputs are listed in Tables \ref{tab:results-designs-70b-inc}-\ref{tab:results-designs-8b-dec} in the Appendix.

\subsection{\model's application of behavioral knowledge in reasoning and problem solving}
\label{sec:infer-knowledge}

Incorporating foundational knowledge from behavioral science is essential for advancing research and applications in the field. 
Built upon a large language model, \model leverages advanced language processing %
to systematically extract, organize, and represent behavioral science knowledge. In general, these tasks can be represented as 
\begin{equation}
    \mathbb{K} \rightarrow \Delta \mathbb{K},
\end{equation}
where $\mathbb{K}$ is the union of the existing knowledge, and the solution to a complex problem results in new piece of knowledge $\Delta \mathbb{K}$ that is derived from $\mathbb{K}$ and can be added into $\mathbb{K}$. 

\begin{table}[t]
\centering
\begin{tabular}{lcccc}
\toprule
 & \multicolumn{2}{c}{\textbf{BLEURT}($\uparrow$)} & \multicolumn{2}{c}{\textbf{ROUGE-1}($\uparrow$)} \\
\textbf{Model} & \textbf{Idea} & \textbf{Title} & \textbf{Idea} & \textbf{Title} \\
\midrule
\model 8B	&	\textbf{0.498}	&	0.438	&	0.076	&	\textbf{0.406}	\\
\model 70B	&	0.423	&	0.427	&	\textbf{0.315}	&	0.385	\\
\midrule
Llama 3.1 8B	&	0.460	&	\textbf{0.441}	&	0.083	&	0.076	\\
Llama 3.1 70B	&	0.482	&	0.398	&	0.106	&	0.238	\\
GPT 4o-mini	&	0.426	&	0.415	&	0.084	&	0.394	\\
GPT 4o	&	0.452	&	0.412	&	0.136	&	0.390	\\
\bottomrule
\end{tabular}
\vspace{5pt}
\caption{
Performance of models on research workflow prediction. 
Given research contexts, the model-generated ideas are compared with the key ideas extracted from publication abstracts. Given the research workflows, the model-generated titles are directly compared with actual paper titles. 
For each prediction scenario, BLEURT and ROUGE-1 scores are reported. Higher scores indicate better predictions. 
\textbf{Bold} texts highlight the best performances. 
}
\label{tab:results-sci-flow}
\end{table}

\paragraph{Research workflow reasoning. }
A behavioral foundation model may represent and apply knowledge to assist behavioral science research. Following recent work that benchmarks LLMs in predicting scientific research workflows \citep{zhang2024massw}, we evaluate \model’s ability to reason research workflows in behavioral science. We focus on two subtasks: (1) generating a key research idea based on background context, and (2) generating a paper title given the context, key idea, method, outcome, along with projected impact. For evaluation, we use a subset of 300 AER publications, held-out from  training. 
Table \ref{tab:results-sci-flow} presents the performance of different models, where \model outperforms baseline LLMs on three out of four metrics, highlighting its potential in assisting behavioral science research. 

\paragraph{Complex problem solving. }
\begin{wraptable}{r}{7cm}
\vspace{-12pt}
\hspace*{16pt}
\begin{minipage}{6cm}
\centering
\begin{tabular}{lc}
    \toprule
    \textbf{Model} & \textbf{Accuracy}(\%) \\
    \midrule
    \model 8B & 51.3\%\\
    \model 70B & \underline{73.3\%}\\
    \midrule    
    Llama 3.1 8B & 48.4\% \\
    Llama 3.1 70B & 68.8\% \\
    GPT 4o-mini & 72.9\% \\
    GPT 4o & \textbf{81.6\%} \\
    \bottomrule
\end{tabular}
\vspace{5pt}
\caption{
Performances on IEO contest problems.  
\textbf{Bold} highlights the best performance, while \underline{underline} highlights the second best. 
}
\label{tab:results-qa}
\end{minipage}
\vspace{-20pt}
\end{wraptable}

Researchers often use questions from competitions like the International Mathematical Olympiad (IMO) to evaluate the reasoning and complex problem-solving abilities of general foundation models such as ChatGPT and Llama. In specialized fields, complex question-answering benchmarks also serve as a metric for assessing a model’s ability to represent accurate, reliable, and contextually appropriate knowledge.

We employ problems from the International Economics Olympiad (IEO)\footnote{Official website of the International Economics Olympiad (IEO): \url{https://ieo-official.org/}, retrieved April 9, 2025.} to test \model's  ability to represent, reason about, and apply domain-specific knowledge.
We curate five years of IEO contests, from 2019 to 2024, collecting 91 multiple-choice questions.

We aggregate the results from 10 independent experiments. 
Table \ref{tab:results-qa} presents the average accuracy of different models in answering IEO questions. Although \model was not explicitly trained on IEO data, we observe a notable improvement over the Llama 3.1 base models (from 48.4\% to 51.3\% for 8B and from 68.8\% to 73.3\% for 70B), indicating that the trained foundation model can better represent and apply behavioral knowledge. 
Among all models, GPT-4o achieves the highest accuracy (81.6\%), likely benefiting from its significantly larger parameter scale (hundreds of billions) and specific training for reasoning. 

\subsection{Summary}

The series of benchmarking tasks show that \model has obtained some generalizable abilities across a range of behavior modeling, prediction, and analysis tasks. Compared with the base LLMs and other commercial LLMs, \model models excel in tasks that are identical to the training tasks (but evaluated with held-out data): e.g., personality trait prediction, research flow reasoning. In individual tasks where sufficient training data is available, an expert model (e.g., a agentic model) that is specifically trained on the same type of data may perform better than a foundational model (See Appendix B). However, their performance is unlikely to be generalizable to other tasks.  \model shows advanced performance on tasks that are related to the training data but deviate from the training tasks: e.g., behavior distribution prediction and demographics prediction,  demonstrating the ability of foundation models to synthesize knowledge from the training data and generalize it to solve different tasks. What is perhaps most interesting is the ability in tasks that are not related to the training data: e.g., context inference, IEO contest. \model presents intriguing generalizability to these unseen and intellectually challenging tasks, although further analyses are needed to better understand the performance. A quantitative evaluation procedure is needed to measure the success of context inference, and more experiments are needed to demonstrate and explain the ability of behavioral foundation models in reasoning and problem-solving tasks as complex as the IEO tests.

\section{Related Work}

\subsection{Foundation Models for Behavioral Science}

In scientific fields that are related to studying human behaviors, general foundation models such as GPT and Llama are being increasingly applied \citep{demszky2023using,meng2024ai,bail2024can}. These models show promise in improving survey research \citep{argyle2023out,grossmann2023ai}, enabling automated content analysis \citep{wu2023large,yang2023large,gilardi2023chatgpt}, supporting agent-based modeling \citep{park2023generative,park2024generative}, and predicting experiment results \citep{hewitt2024predicting,lippert2024can}.
However, recent work highlights significant challenges and limitations in directly applying general-purpose models to tasks of more importance to behavioral science. For instance, these models often fail to capture population variance \citep{bisbee2024synthetic}, exhibit biases toward specific data formats \citep{dentella2023systematic}, and underrepresent certain population groups \citep{santurkar2023whose}. This gap between the needs of behavioral science and the design of general foundation models underscores the demand for domain-specific models that align with the principles and data of behavioral science.

Despite this need, behavior-related foundation models remain largely underexplored, with only a few relevant efforts. For instance, \citet{binz2024centaur} introduce Centaur, a cognitive science foundation model trained on psychological experiment data and tested for predicting human responses in cognitive experiments such as decision-making, memory, supervised learning, and Markov decision processes.
Another example is Monad, a private model trained on behavioral event data \citep{rychalska2023synerise}. Trained with user behavioral events such as clicks, likes, page views, card transactions, add-to-basket,
or purchases, the main focus of Monad is to produce behavioral representations as large vectors that capture the behavioral patterns of users. 
Another line of work includes Meta Motivo and related models \cite{tirinzoni2025zero, sikchi2025fast}, which capture human behavior through full-body physical motion. These models can control physics-based virtual humanoid agents across a wide range of whole-body tasks. However, they do not address higher-level decision-making processes involved in complex scenarios such as economic activities.

Our experiments demonstrate that \model supports a variety of behavioral science tasks, whereas existing models like Centaur and Monad specialize in cognitive response prediction and user behavior representation.

\subsection{AI Behavioral Science}

\emph{AI behavioral science} \citep{jackson2025AI} is inherently bidirectional--not only AI can serve as a tool in behavioral science research as discussed above, but also behavioral science methods can be used to study the behaviors, biases, and emergent properties of AI systems themselves.
For example, \citet{aher2023using,mei2024turing} and \cite{xie2024different} have conducted variations on  Turing tests, comparing LLMs' decisions in a  economic games to human data. Similarly, \citet{horton2023large, lore2024strategic} investigated the strategic behavior of LLMs, while \citet{chen2023emergence} assessed GPT’s economic rationality. Other studies applied behavioral frameworks such as theory of mind to LLMs \citep{strachan2024testing}, or examined biases in AI-generated survey responses \citep{salecha2024large}.

While these two directions--using AI for behavioral science and applying behavioral science to AI--have each advanced, a unifying approach remains missing. Specifically, there is a lack of models that function both as tools for understanding human behavior and as platforms for behavioral interrogation of AI systems themselves.
Our work bridges this gap by introducing a foundation model particularly for behavioral science. %

\section{Conclusion and Discussion}
\model is a family of open foundation models specifically designed for modeling human behaviors and solving tasks in behavioral science. Trained on a carefully curated corpus encompassing a diverse range of behavioral data, we demonstrate \model's ability in:
(1) Predicting and simulating behavior across diverse scenarios;
(2) Inferring subject characteristics at both individual and population levels; 
(3) Inferring contextual factors influencing behavior; and 
(4) Applying behavioral science knowledge in IEO problem-solving and research workflow prediction. 

Our work aims to establish a systematic framework for training and evaluating behavioral foundation models, and to offer guidance for future directions in their optimization. We plan to expand the training corpus with additional datasets and tasks that represent diverse instantiations of Equation \ref{eqn:framework}. One may also explore alternative base LLMs beyond those used in the current version. The initial versions of \model are purely trained using supervised fine-tuning, without further reinforcement learning based training, which has been shown to be critical to improve the reasoning abilities of foundation models. To demonstrate the generalizable abilities of \model on multiple tasks, we intentionally did not use reinforcement learning to optimize the performance in particular tasks. RL-based reasoning will be equipped in future versions of \model. We envision \model as an initial step toward fostering continuous engagement from the broader community to develop behavioral foundation models that serve as: 

\paragraph{A new interface for behavioral science research. }
Behavioral foundation models introduce a novel interface for researchers, bridging traditional research methodologies with AI-driven insights. Unlike conventional task-specific models, \model provides a unified and scalable framework for representing, predicting, and reasoning about human behavior across diverse contexts. Along this direction, behavioral foundation models are anticipated to further facilitate scientist-AI collaboration, enabling researchers to generate hypotheses, design experiments, run in silico experiments, and analyze data in ways that go beyond simple automation. %

\paragraph{A new foundation for applications across various domains. }
We anticipate the impact of behavioral foundation models to extend across multiple domains. Social scientists can leverage their advanced analytical capabilities to study complex social dynamics, uncover behavioral patterns, design and conduct in silico experiments at scale and at low cost. Engineers and AI developers can integrate behavioral insights into system design, enhancing human-AI interactions and adaptive decision-making. Industries reliant on behavioral analytics and A/B testing can use \model to optimize decision-making, refine market strategies, and exploring interventions to improve user experiences. %

\newpage
\bibliographystyle{plainnat}  
\bibliography{references}  %

\newpage
\appendix
\section{Model Configuration and Training}
\label{app:train}

\subsection{Backbone Foundation Model and Implementation Details}

\textbf{Model.}
We used \texttt{LLaMA-3.1-8B-Instruct}\footnote{\url{https://huggingface.co/meta-llama/Llama-3.1-8B-Instruct}, retrieved April 9, 2025. } and \texttt{LLaMA-3.1-70B-Instruct}\footnote{\url{https://huggingface.co/meta-llama/Llama-3.1-70B-Instruct}, retrieved April 9, 2025. } as backbone models. The 70B variant was quantized to 8-bit precision using the \texttt{bitsandbytes}\footnote{\url{https://huggingface.co/docs/bitsandbytes}, retrieved April 9, 2025. } library.

\textbf{Method.}
Fine-tuning was conducted using LlamaFactory\footnote{\url{https://github.com/hiyouga/LLaMA-Factory}, retrieved April 9, 2025.} \citep{zheng2024llamafactory} with supervised fine-tuning (SFT). Low-Rank Adaptation (LoRA) \citep{hu2022lora} was applied across all model layers to enable efficient fine-tuning with reduced memory overhead.

\textbf{Training Configuration and Hyperparameters.}
Training was conducted with a per-device batch size of 1 and 8 gradient accumulation steps, using a learning rate of $10^{-4}$ with a cosine scheduler and a warmup ratio of 0.1. The model was trained for 3 epochs using \texttt{bf16} mixed precision. A distributed data parallel (DDP) timeout of 180,000,000 seconds was set. This configuration balanced efficiency and convergence stability for large-scale fine-tuning.

\textbf{Data Formatting.}
Data were formatted according to the Alpaca template \citep{taori2023alpaca}, aligning with the model’s expected input structure. Each training instance consisted of an \texttt{instruction}, \texttt{input}, and the corresponding ground-truth \texttt{output}. Preprocessing steps included tokenization, truncation, caching, and multiprocessing to enhance throughput and minimize latency.

\subsection{Training Data and Training Tasks}

\subsubsection{Literature Data}
\label{app:data-literature}

\paragraph{Publications. }

Our dataset comprises 3,003 publications from the leading journal, \emph{American Economic Review} (AER), which publish interdisciplinary research spanning economics, psychology, sociology, and human behavior. 
Articles were selected based on the availability of both title and abstract, which are essential for model training. Of the total, 2,703 publications are used for training and the remainder for evaluation.

\paragraph{Research workflow reasoning. } 

We convert publication data into a research workflow reasoning format. Following \citet{zhang2024massw}, we extract five key components from each paper--context, key idea, method, outcome, and projected impact--using the paper's title and abstract. Extraction is performed using \texttt{gpt-4o-2024-05-01-preview}. The data are then structured into two subtasks: (1) idea generation, and (2) title prediction. 

For idea generation, the \texttt{input} consists of the extracted context (the template is provided below), and the \texttt{output} is the corresponding key idea. 

\begin{framed}
Given the context: '\{context\}', generate key ideas that could advance this area of study.
\end{framed}

For title prediction, the \texttt{input} includes all five extracted components (the template is provided below), with the original paper title as the \texttt{output}. These input-output pairs are then used to train the model.

\begin{framed}
Given the context: '\{context\}', the key idea: '\{key\_idea\}', the method: '\{method\}', the
outcome: '\{outcome\}', and the future impact: '\{future\_impact\}', predict the title of this
research paper. The title should be concise and reflective of the core aspects.
\end{framed}

The \texttt{instruction} is given as follows:
\begin{framed}
You are an expert in research tasked with generating detailed prompts for various aspects of academic research papers.
Each task involves creating a specific type of prompt based on the provided information. Here are the definitions of each part you will work with:

- Concept: Definition, Relative Time

- Context: The status quo of related literature or reality which motivated this study.
This could normally be a problem, a research question, or a research gap that has not been successfully addressed by previous work. This is anything that happened before this study.

- Key Idea: The main intellectual merit of this paper, often in comparison to the context.
This could normally be a novel idea or solution proposed in this paper that distinguishes it from what's already done in literature.
This is proposed in this study.

- Method: The specific research method that investigates and validates the key idea.
This could be an experimental setup, a theoretical framework, or other necessary methodology to implement and/or evaluate the key idea.
This is performed in this study.

- Outcome: The factual statement about the study output.
This could be the experiment results and any other measurable outcome that has occurred.
It marks whether the key hypothesis is testified or not. This is produced in this study.

- Projected Impact: The author-anticipated impact of the work on the field, and potential further research identified by the author that may improve or extend this study.
This is anything being anticipated but has not happened yet.
\end{framed}

\subsubsection{Experimental Data}
\label{app:data-experiment}

\paragraph{MobLab human-play records. }

We use data from MobLab\footnote{MobLab: \url{https://moblab.com/}, retrieved April 11, 2025.}, a platform for behavioral science and economic games. Our analysis focuses on five classic economic games across seven roles: (i) Dictator Game, (ii) Ultimatum Game (Proposer and Responder), (iii) Trust Game (Investor and Banker), (iv) Public Goods Game, and (v) Bomb Game. Additional background on these games is provided in \citet{mei2024turing}. The dataset includes first-round gameplay data from 68,779 subjects, yielding 82,057 independent observations collected between 2015 and 2023.
Since the evaluation regarding this data collection is performed at the distributional level, we do not separate the data into training and test sets.

\paragraph{Behavior distribution prediction. }

To train the model to predict behavioral distributions, we format each human-play record as a single data entry. The game instruction (as described in \citet{mei2024turing}) is provided as the \texttt{instruction}, and the observed human behavior is recorded as the \texttt{output} (e.g., ``[\$50]’’ in the Dictator Game). In total, we construct 82,057 such data entries for model training.

\subsubsection{Survey Data}
\label{app:data-survey}

\paragraph{BigFive personality test. }

We use the publicly available OCEAN Five Factor Personality Test Responses dataset\footnote{\url{https://www.kaggle.com/datasets/lucasgreenwell/ocean-five-factor-personality-test-responses}, retrieved April 11, 2025.}, sourced from the Open-Source Psychometrics Project\footnote{\url{https://openpsychometrics.org/}, retrieved April 11, 2025.}, a nonprofit initiative dedicated to public education and psychological research.
The dataset includes responses and metadata from 19,719 subjects. Participants represent a wide range of demographic backgrounds, covering individuals from over 11 racial groups and 161 countries and regions. Ages range from 13 and above, providing broad age coverage. Gender identity is self-reported, with participants identifying as male, female, or other, reflecting gender diversity in the sample.
Of the subjects, 17,667 are used for training, and 1,963 are reserved for evaluation.

\paragraph{BigFive personality traits prediction. }

To train the model to predict survey responses, each data entry is formatted as follows: for each subject and each personality dimension, we provide the subject’s demographic information in the \texttt{instruction}, the prediction target in the \texttt{input}, and the ground-truth dimensional score (an integer from 10 to 50) in the \texttt{output}. The \texttt{instruction} and \texttt{input} are templated as follows:

\begin{framed}
\texttt{instruction}:\newline
You are a helpful AI assistant. The user has the following demographics: \{demographics\}.

\texttt{input}:\newline
Based on this person's demographics, rate their \{personality\_dimension\} on a scale of [10] to [50]. 

For openness:
[10] means very conventional/traditional, [50] means very open to new experiences.\newline
For conscientiousness:
[10] means very spontaneous/flexible, [50] means very organized/responsible.\newline
For extroversion:
[10] means very introverted, [50] means very extroverted. \newline
For agreeableness:
[10] means very competitive/assertive, [50] means very cooperative/compassionate.\newline
For neuroticism:
[10] means very emotionally stable, [50] means very prone to stress/anxiety.\newline

The response should only be a number from 10 to 50 in square brackets.
\end{framed}

\paragraph{Demographics prediction. }

Another modeling direction involves predicting demographic attributes from survey responses. Each subject’s record is converted into a data entry, where the overall task description is provided in the \texttt{instruction}, the subject’s personality scores in the \texttt{input}, and the subject’s age (an integer) as the ground-truth \texttt{output}. The \texttt{instruction} and \texttt{input} are templated as follows:

\begin{framed}
\texttt{instruction}: \newline
You are an expert in psychology. Given a person's BigFive personality scores, your task is to predict this person's demographics. \newline

\texttt{input}: \newline
\#\# BigFive Dimensionality Scores\newline
Each dimensionality score ranges from 10 to 50, with 10 indicating the lowest score in that dimension and 50 indicating the highest score. \newline
Openness:  \{openness\_score\} \newline
Conscientiousness: \{conscientiousness\_score\} \newline
Extroversion: \{extroversion\_score\} \newline
Agreeableness: \{agreeableness\_score\} \newline
Neuroticism: \{neuroticism\_score\} \newline\newline
\#\# Output Format \newline
Please predict this person's gender. Choose from the following: 1=Male, 2=Female, 3=Other. Please only output a single number indicating the choice highlighted with [] (e.g., [x]).
\end{framed}

\section{Model Evaluation}
\label{app:benchmarks}

\subsection{Inferring Behaviors}

\paragraph{Predict behavior distribution in economic games. }

Using the human-play data described in Appendix \ref{app:data-experiment}, we evaluate how \model performs in predicting behavioral distributions in economic games. For each game, we generate behaviors from 1,000 independent chat sessions.
During generation, the game instructions (as provided in \citet{mei2024turing}) are given as user prompts, with system prompts left blank.
The generated behaviors are then compared with human play data from the MobLab platform. 
Performance is assessed at the distributional level using the Wasserstein distance between the observed and simulated distributions. LLaMA 3.1 models and GPT models (\texttt{gpt-4o-2024-08-01} and \texttt{gpt-4o-mini-2024-07-18}) are included as baselines following the same procedure. Evaluation results are reported in Table \ref{tab:results-behavior-moblab}.

\paragraph{Predict Big Five personality traits. }

To predict subjects’ personality traits as assessed by the Big Five survey, we follow the data formatting described in Appendix \ref{app:data-survey}. Demographic information (including age, gender, nationality, race, etc.) is provided, and models are tasked with predicting scores for the five personality dimensions: openness, conscientiousness, extroversion, agreeableness, and neuroticism. Evaluation is conducted on the held-out set of 1,963 subjects to ensure no data leakage.
We report both individual-level metrics (mean absolute error and Spearman's rank correlation) and distributional alignment via the Wasserstein distance between predicted and ground-truth score distributions. In addition to general-purpose LLM baselines (LLaMA 3.1 and GPT models), we compare against expert models fine-tuned specifically for this task (\model 8B$^*$ and \model 70B$^*$), as well as an XGBoost regressor. Full results are presented in Table \ref{tab:results-behavior-bigfive-full}.

\begin{table}[htbp]
\centering
\begin{tabular}{lccccc} 
\toprule
\textbf{Model} & \textbf{MAE}($\downarrow$)  & \textbf{Spearman's corr}($\uparrow$) & \textbf{W-distance}($\downarrow$) \\ 
\midrule
\model 8B                      & 7.27 & 0.101$^\ddagger$    & 1.75        \\
\model 70B                     & 7.27 & 0.128$^\ddagger$    & 1.68$^\dagger$        \\
\midrule
Llama 3.1 8B                   & 8.23 & 0.037   & 5.26        \\
Llama 3.1 70B                  & 7.05 & 0.029   & 4.06        \\
GPT 4o-mini                    & 6.85 & 0.034   & 5.52        \\
GPT 4o                         & 7.11 & 0.029   & 5.00        \\
\midrule
\model 8B$^*$                  & 7.29 & 0.098$^\ddagger$    & 1.71        \\
\model 70B$^*$                 & 7.22 & 0.109$^\ddagger$    & 1.93        \\
XGBoost                        & 6.03 & 0.252$^\ddagger$    & 4.59        \\
\bottomrule
\end{tabular}
\vspace{5pt}
\caption{
Performance on predicting Big Five personality scores given demographics. For each model, the mean absolute error (MAE), Spearman's correlation coefficient, and Wasserstein distance are computed by comparing the predictions and ground truths. Metrics are aggregated over five personality dimensions (OCEAN). 
``$^*$'' indicates \model models fine-tuned only on this task. XGBoost indicates an expert model trained directly through the behavioral data without using a LLM as backbone. 
$^\ddagger$ indicates that the correlation is significant ($p<0.05$). 
``$^\dagger$'' indicates the distributions that pass the smoothed Kolmogorov–Smirnov test ($p>0.05$) with a bin width of 10. 
}
\label{tab:results-behavior-bigfive-full}
\end{table}

\subsection{Inferring Subjects/Populations}

\paragraph{Demographics prediction. }

Using the data formatting described in Appendix \ref{app:data-survey}, we predict subjects’ demographic attributes, specifically age, based on their personality dimension scores. Evaluation includes both individual level metrics (mean absolute error and Spearman’s rank correlation) and distributional alignment, measured by the Wasserstein distance between predicted and ground truth age distributions. Full results are reported in Table \ref{tab:results-subjects-bigfive-full}.

\begin{table}[htbp]
\centering
\begin{tabular}{lcccc}
\toprule
\textbf{Model} & \textbf{MAE}($\downarrow$)  & \textbf{Spearman's corr}($\uparrow$) & \textbf{W-distance}($\downarrow$) \\ 
\midrule
\model 8B                      & 9.05  & 0.108$^{\ddagger}$     & 5.39       \\
\model 70B                     & 9.16  & 0.134$^{\ddagger}$     & 5.70       \\
\midrule
Llama 3.1 8B                   & 10.50 & 0.038    & 8.62       \\
Llama 3.1 70B                  & 10.36 & 0.029    & 7.75       \\
GPT 4o-mini                    & 10.17 & 0.030    & 8.07       \\
GPT 4o                         & 11.01 & 0.048$^{\ddagger}$    & 9.12       \\
\midrule
\model 8B$^*$                  & 9.21  & 0.096$^{\ddagger}$     & 5.56       \\
\model 70B$^*$                 & 9.35  & 0.107$^{\ddagger}$     & 5.66       \\
XGBoost                        & 8.97  & 0.168$^{\ddagger}$    & 5.03       \\
\bottomrule

\end{tabular}
\vspace{5pt}
\caption{
Performances on inferring subjects' ages given Big Five personality scores. 
For each model, the mean absolute error (MAE), Spearman's correlation coefficient, and Wasserstein distance are computed by comparing the age prediction results and ground truths. 
``$^*$'' indicates \model models fine-tuned only on this task. XGBoost indicates an expert model trained directly through the behavioral data without using a LLM as backbone. 
$^\ddagger$ indicates that the correlation is significant ($p<0.05$).}
\label{tab:results-subjects-bigfive-full}
\end{table}

\subsection{Inferring Contexts}

\paragraph{Experiment design prediction. }

We conduct a qualitative analysis to evaluate \model’s ability to generate insights about contextual factors influencing behavior. Specifically, given a change in behavior distribution, we assess whether \model can infer plausible situational conditions or experimental settings that might have caused the shift. For this analysis, we reference a meta-study of the Dictator Game \citep{engel2011dictator}, which catalogs a range of experimental designs and their associated outcomes.

We task \model with proposing treatments that could increase/decrease sharing behavior in the Dictator Game, using the following prompt:

\begin{framed}
You are an expert in behavioral economics. Given the observation of an experiment on the Dictator game, your task is to infer what experiment designs could lead to the observed treatment effect. In the Dictator game, given an endowment of money, one player (the dictator) chooses how much of the money to keep and how much to give to a second player. \newline
In an economic experiment of the Dictator game, we observed the subject behaviors -- the proportion of money to share -- increased/decreased compared to the standard game design. Please list possible experiment designs that could lead to this treatment effect. Please rank output experiment designs by confidence. 
\end{framed}

The model outputs are listed in Tables \ref{tab:results-designs-70b-inc}-\ref{tab:results-designs-8b-dec} and discussed in Sec. \ref{sec:infer-context}. 

\begin{table}[htbp]
\tiny
\centering
\begin{tabular}{ll}
\toprule
\textbf{Run} & \textbf{List of Outputted Experiment Designs} \\
\midrule
1 & \makecell[{{p{15cm}}}]{
(1) Social Identity Priming – Emphasizing group membership or common goals to foster a sense of connection. (Confidence: 50\%) \newline
(2) Emotional Appeal – Using stories or emotional framing to increase empathy and generosity. (Confidence: 45\%) \newline
(3) Transparency \& Accountability – Making decisions public or adding accountability measures to encourage sharing. (Confidence: 38\%) \newline
(4) Incentivizing Altruism – Offering rewards or recognition for generous behavior. (Confidence: 35\%) \newline
(5) Fairness Calibration – Adjusting perceptions of fairness through feedback or endowment modifications. (Confidence: 31\%) \newline
(6) Reciprocity Framing – Suggesting future reciprocation possibilities to encourage sharing. (Confidence: 27\%) \newline
(7) Moral Nudges – Using ethical prompts to remind participants of social and moral expectations. (Confidence: 24\%) \newline
(8) Role-Reversal Perspective – Encouraging dictators to imagine themselves in the recipient’s position. (Confidence: 21\%) \newline
(9) Cultural Context Influence – Testing generosity levels across different cultural norms and values. (Confidence: 17\%) \newline
(10) Temporal Dynamics – Modifying time constraints or introducing future dependence on generosity. (Confidence: 14\%)
} \\
\midrule
2 & \makecell[{{p{15cm}}}]{
(1) Social Identity Priming – Emphasizing shared group membership or common goals to encourage generosity. \newline
(2) Emotional Appeals through Storytelling – Using narratives to engage emotions and increase empathy-driven sharing. \newline
(3) Transparency \& Accountability Mechanisms – Making decisions public or increasing scrutiny to encourage fairer distribution. \newline
(4) Fairness \& Equity Framing – Reframing the decision to highlight fairness and moral responsibility. \newline
(5) Incentivizing Altruism – Offering rewards or recognition to reinforce generous behavior. \newline
(6) Role-Reversal Exercises – Encouraging dictators to consider the recipient's perspective to foster empathy. \newline
(7) Cultural \& Social Norm Interventions – Introducing cultural or societal norms that promote sharing. \newline
(8) Feedback on Social Impact – Providing information on how sharing benefits the recipient and society.
} \\
\midrule
3 & \makecell[{{p{15cm}}}]{
(1) Social Identity Influence – Emphasizing in-group affiliations to encourage sharing. \newline
(2) Emotional Appeal Interventions – Using storytelling or imagery to evoke empathy and generosity. \newline
(3) Transparency \& Accountability – Publicly disclosing decisions to encourage socially responsible behavior. \newline
(4) Default Option Nudges – Setting a higher default sharing percentage to leverage status quo bias. \newline
(5) Incentivizing Altruism – Providing rewards or recognition for generous behavior. \newline
(6) Feedback Mechanisms – Giving immediate feedback on the impact of sharing to reinforce generosity. \newline
(7) Cultural \& Contextual Factors – Introducing communal values to align sharing behavior with cultural norms.
} \\
\midrule
4 & \makecell[{{p{15cm}}}]{
(1) Social Identity Influence – Emphasizing a shared group identity between the dictator and recipient to encourage in-group favoritism and greater sharing. \newline
(2) Transparency \& Accountability – Making the dictator’s decision public to encourage socially responsible behavior and avoid disapproval. \newline
(3) Emotional Connection – Creating a personal interaction between the dictator and recipient to increase empathy and generosity. \newline
(4) Fairness Priming – Exposing the dictator to fairness-related concepts before the game to subtly nudge them toward more equal sharing. \newline
(5) Recipient’s Feedback Mechanism – Allowing recipients to rate or provide feedback on the dictator’s decision to encourage generosity. \newline
(6) Charitable Context – Framing the decision as contributing to a charity or a needy individual to enhance altruism. \newline
(7) Incentivizing Altruism – Offering bonuses or rewards for sharing a higher proportion of the money. \newline
(8) Long-term Interaction – Structuring the game to include repeated interactions, fostering trust and cooperative behavior. \newline
(9) Cultural \& Social Norms – Conducting experiments in different cultural contexts or emphasizing norms that promote generosity. \newline
(10) Cognitive Load Manipulation – Reducing decision complexity to encourage intuitive and generous behavior.
} \\
\midrule
5 & \makecell[{{p{15cm}}}]{
(1) Social Identity Priming – Emphasizing the dictator’s social identity or group affiliation to promote fairness and social harmony. (High confidence) \newline
(2) Charity Frame – Reframing the recipient as a charitable cause to activate altruistic tendencies. (High confidence) \newline
(3) Transparency \& Accountability – Making decisions public to encourage sharing due to social pressure and reputation concerns. (Medium confidence) \newline
(4) Emotional Appeals – Using storytelling or visuals to evoke empathy and increase generosity. (Medium confidence) \newline
(5) Cultural Contexts – Conducting experiments in collectivist cultures to assess natural variations in generosity. (Medium confidence) \newline
(6) Temporal Distance – Emphasizing the immediate impact of sharing to reduce impulsivity and promote generosity. (Low confidence) \newline
(7) Feedback Mechanism – Providing post-experiment feedback on the impact of sharing decisions to encourage altruism. (Low confidence) \newline
(8) Endowment Variability – Varying the amount or source of the endowment to study entitlement effects on sharing behavior. (Low confidence)
} \\
\bottomrule
\end{tabular}
\caption{Experiment designs outputted by \model 70B when asked to propose treatments that could increase sharing behavior in the Dictator Game.}
\label{tab:results-designs-70b-inc}
\end{table}

\begin{table}[htbp]
\tiny
\centering
\begin{tabular}{ll}
\toprule
\textbf{Run} & \textbf{List of Outputted Experiment Designs} \\
\midrule
1 & \makecell[{{p{15cm}}}]{
(1) Social Identity Influence – Emphasizing a shared identity may reduce sharing due to perceived competition within the group. \newline
(2) Anonymity Variation – Reducing anonymity might lead to lower sharing as dictators manage their self-image and avoid scrutiny. \newline
(3) Cognitive Load Manipulation – Increasing cognitive burden can reduce consideration for fairness and generosity. \newline
(4) Temporal Distance – Delaying the transfer of money may lead to lower sharing due to discounting future consequences. \newline
(5) Emotional State Manipulation – Inducing negative emotions (e.g., stress, frustration) may decrease empathy and sharing. \newline
(6) Incentive Structures – Introducing rewards for low sharing or penalties for high sharing can directly discourage generosity. \newline
(7) Feedback Mechanism – Providing feedback that past generosity was unnoticed or unvalued may discourage future sharing. \newline
(8) Recipient Characteristics – Framing the recipient as less deserving or in less need can reduce altruistic behavior. \newline
(9) Contextual Framing – Emphasizing financial scarcity or personal gain may shift focus away from generosity. \newline
(10) Cultural Context – Conducting the experiment in an individualistic culture may naturally result in lower sharing due to differing social norms.
} \\
\midrule
2 & \makecell[{{p{15cm}}}]{
(1) Social Comparison \& Anchoring – Informing the dictator of past average donations may anchor decisions toward lower sharing. \newline
(2) Time Delay in Decision Announcement – Allowing reflection time may reduce impulsive generosity and lower sharing. \newline
(3) Mandatory Donation Constraints – Imposing a minimum or maximum donation may limit perceived autonomy and decrease sharing. \newline
(4) Third-Party Observation – Informing the dictator that their decision will be reviewed may lead to reduced sharing due to accountability concerns. \newline
(5) Endowment Source Modification – Dictators who earn their endowment may feel more entitled to keep it, decreasing sharing. \newline
(6) Loss-Framing Effect – Emphasizing the "loss" from giving rather than the "gain" from keeping may reduce generosity. \newline
(7) Recipient Anonymity Variation – Making the recipient’s identity known could trigger biases or prejudices, reducing sharing. \newline
(8) Personal Cost of Donation – Introducing effort or cost into the sharing decision may deter generosity. \newline
(9) Varying Endowment Size – Larger endowments may lead to a lower proportion shared due to psychological factors. \newline
(10) Feedback on Donation Impact – Receiving information about the effect of sharing may make dictators feel their contribution is unnecessary. \newline
(11) Virtual vs. Real Money – Using hypothetical currency may lead to lower generosity since the decision carries less perceived weight. \newline
(12) Contextual Stress Factors – Introducing external stressors, such as financial downturns, may increase self-interest and reduce sharing.
} \\
\midrule
3 & \makecell[{{p{15cm}}}]{
(1) Anonymity Removal – Making the dictator’s identity and decision public may lead to strategic selfishness, reducing sharing. \newline
(2) Social Norms Priming – Priming competitive or selfish norms before the game may decrease the proportion of money shared. \newline
(3) Recipient Feedback Mechanism – Allowing recipients to express dissatisfaction may discourage dictators from sharing in subsequent rounds. \newline
(4) Temporal Delay in Decision-Making – Introducing a delay before allocation may encourage reflection and lead to more selfish decisions. \newline
(5) External Incentives for Selfishness – Providing rewards for keeping a larger portion of the endowment may directly reduce sharing. \newline
(6) Role-Reversal Awareness – If framed as a threat, awareness of future role reversals may trigger risk aversion, leading to lower generosity. \newline
(7) Emotional Manipulation – Inducing stress or anxiety in the dictator may result in more conservative, self-focused decisions.
} \\
\midrule
4 & \makecell[{{p{15cm}}}]{
(1) Identity Disclosure – Requiring dictators to reveal their identity may trigger social scrutiny, reducing generosity. (Confidence: 50\%) \newline
(2) Social Comparison – Showing past participants' average donations may encourage lower sharing by normalizing selfishness. (Confidence: 40\%) \newline
(3) Emphasizing Personal Gain – Using visual or textual emphasis on the benefits of keeping money may shift focus toward self-interest. (Confidence: 30\%) \newline
(4) Time Constraints – Implementing a short decision window may push dictators toward instinctive, self-preserving choices. (Confidence: 20\%) \newline
(5) Loss Framing – Framing sharing as a loss rather than a gain may activate loss aversion and reduce generosity. (Confidence: 10\%)
} \\
\midrule
5 & \makecell[{{p{15cm}}}]{
(1) Social Identity Component – Introducing shared group identity may activate biases or reduce sharing toward out-group members. \newline
(2) Increasing Endowment Size – Larger endowments may lower the proportion shared due to perceived greater loss or reduced impact. \newline
(3) Decision-Making Delay – A time delay before money transfer may allow more self-interested, rationalized decisions. \newline
(4) Reciprocity Norms Introduction – Suggesting reciprocity may lead dictators to share less, assuming giving is sufficient without a large contribution. \newline
(5) Enhanced Anonymity – Greater anonymity can reduce social accountability, encouraging more selfish decisions. \newline
(6) Feedback on Recipient Satisfaction – Learning that recipients are satisfied with smaller amounts may reinforce lower sharing. \newline
(7) Self-Interest Priming – Framing the game with self-interest cues may nudge dictators toward prioritizing personal gain. \newline
(8) Recipient Status Framing – Highlighting the recipient’s resources may make them seem less in need, reducing sharing. \newline
(9) Non-Monetary Incentive Structure – Using points or badges instead of money may alter perceived value and decrease generosity. \newline
(10) Decision Environment Change – Making the setting more formal or complex may constrain sharing by emphasizing structure over fairness.
} \\
\bottomrule
\end{tabular}
\caption{Experiment designs outputted by \model 70B when asked to propose treatments that could decrease sharing behavior in the Dictator Game.}
\label{tab:results-designs-70b-dec}
\end{table}

\begin{table}[htbp]
\tiny
\centering
\begin{tabular}{ll}
\toprule
\textbf{Run} & \textbf{List of Outputted Experiment Designs} \\
\midrule
1 & \makecell[{{p{15cm}}}]{
(1) Larger Endowment, Social Norms \& Feedback – Increase endowment size, introduce norms about average sharing behavior, and provide feedback on past decisions. (Confidence: 0.9) \newline
(2) Incentive Structure \& Identity Disclosure – Reward higher sharing, allow participants to self-identify as "generous" or "frugal," and inform them of recipient preferences. (Confidence: 0.7) \newline
(3) Group Decision-Making \& Emotional Appeals – Allow participants to collectively decide on sharing, use emotional framing, and introduce role-playing. (Confidence: 0.4) \newline
(4) Loss Aversion \& Cognitive Biases – Frame decisions as losses versus gains and leverage biases like the endowment effect. (Confidence: 0.2)
} \\
\midrule
2 & \makecell[{{p{15cm}}}]{
(1) Modified Payoff Structure – Introduce a bonus/penalty system to incentivize sharing. (Highest confidence) \newline
(2) Social Identity Manipulation – Emphasize social identity to increase group loyalty and generosity. (High confidence) \newline
(3) Real-World Context Framing – Frame the game as having real-world consequences, enhancing empathy. (High confidence) \newline
(4) Feedback Mechanism – Provide feedback on the impact of sharing decisions to influence behavior. \newline
(5) Group Decision-Making – Allow groups to decide collectively, leveraging social influence. \newline
(6) Emotional Priming – Use images or scenarios to evoke generosity. \newline
(7) Reciprocity Dynamics – Introduce the possibility of future reciprocation to encourage sharing. \newline
(8) Cultural \& Ethical Contexts – Vary cultural and ethical settings to explore their influence on sharing behavior.
} \\
\midrule
3 & \makecell[{{p{15cm}}}]{
(1) Social Norms \& Expectations – Introduce social norms to encourage conformity and increase sharing. (High confidence) \newline
(2) Visual Representation – Use graphical aids to facilitate decision-making and promote generosity. (High confidence) \newline
(3) Time Constraint – Implement a time limit, as rapid decisions often lead to more generous allocations. (Medium confidence) \newline
(4) Framing Effect – Frame decisions in terms of gains rather than losses to influence generosity. (Medium confidence) \newline
(5) Altering Endowment Amount – Change the initial amount given to the dictator to explore psychological effects. (Low confidence)
} \\
\midrule
4 & \makecell[{{p{15cm}}}]{
(1) Incentive Structures – Introduce bonuses for generous sharing to motivate higher contributions. (High confidence) \newline
(2) Social Identity – Link the dictator to a group emphasizing fairness or altruism to encourage sharing. (High confidence) \newline
(3) Feedback Mechanisms – Inform players about the impact of their sharing decisions on the recipient. (Medium confidence) \newline
(4) Game Context – Make the second player more relatable using avatars or background stories. (Medium confidence) \newline
(5) Loss Aversion – Frame the decision as avoiding a loss rather than achieving a gain. (Medium confidence) \newline
(6) Cultural Influences – Test the effect of cultural norms on sharing behavior. (Low confidence)
} \\
\midrule
5 & \makecell[{{p{15cm}}}]{
(1) Increased Endowment Size – Providing a larger sum may encourage more generosity. (High confidence) \newline
(2) Social Identity Manipulation – Adding recipient context (e.g., backstory, visual cue) to enhance connection. (High confidence) \newline
(3) Incentive Structure Modification – Implementing penalties for low sharing or bonuses for high sharing. (Medium confidence) \newline
(4) Feedback Mechanism Introduction – Providing real-time feedback on sharing decisions to influence behavior. (Medium confidence) \newline
(5) Cultural \& Demographic Factors – Examining variations based on participants’ cultural and demographic backgrounds. (Low confidence)
} \\
\bottomrule
\end{tabular}
\caption{Experiment designs outputted by \model 8B when asked to propose treatments that could increase sharing behavior in the Dictator Game.}
\label{tab:results-designs-8b-inc}
\end{table}

\begin{table}[htbp]
\tiny
\centering
\begin{tabular}{ll}
\toprule
\textbf{Run} & \textbf{List of Outputted Experiment Designs} \\
\midrule
1 & \makecell[{{p{15cm}}}]{
(1) Social Norm Emphasis – Introduce statements about fairness and equity to influence sharing behavior. (High confidence) \newline
(2) Anonymity \& Privacy – Increase anonymity, which may reduce generosity as players feel less accountable. (Medium confidence) \newline
(3) Recipient Framing – Present the recipient as a charity or community instead of an individual, altering perception. (Low confidence) \newline
(4) Endowment Size Variation – Test whether a larger endowment results in the dictator keeping a higher proportion. (Low confidence) \newline
(5) Mirror Image Game – Introduce a reversed role scenario to explore its influence on sharing behavior. (Low confidence)
} \\
\midrule
2 & \makecell[{{p{15cm}}}]{
(1) Increased Endowment Size – A larger sum may lead to less sharing as dictators keep more for themselves. (High confidence) \newline
(2) Incentive Structures – Introducing bonuses or penalties may shift decision-making towards lower sharing. (High confidence) \newline
(3) Social Identity Manipulation – Framing the recipient as an in-group or out-group member may reduce generosity. (High confidence) \newline
(4) Cognitive Load Manipulation – Increasing complexity in decision-making could lead to decision fatigue, reducing sharing. (Medium confidence) \newline
(5) Emotional Priming – Using words or images to influence mood may lead to less sharing. (Medium confidence) \newline
(6) Feedback Mechanism Alteration – Changing the nature or timing of feedback could impact sharing behavior. (Low confidence) \newline
(7) Group Decision-Making – Collective decision-making might alter individual choices, potentially decreasing sharing. (Low confidence)
} \\
\midrule
3 & \makecell[{{p{15cm}}}]{
(1) Cost-Sharing Mechanism – Requiring the dictator to pay a participation fee or penalty may reduce sharing. (High confidence) \newline
(2) Dynamic Endowment Adjustment – Informing players that their endowment is temporary or variable may reduce generosity due to perceived scarcity. (High confidence) \newline
(3) Social Influence \& Peer Comparison – Showing dictators the sharing behavior of others may lead to reduced generosity. (High confidence) \newline
(4) Emotional Priming \& Context Manipulation – Introducing stress, anxiety, or emphasizing individualism may lower sharing. (Medium confidence) \newline
(5) Temporal Discounting Manipulation – Offering the endowment in installments could prioritize immediate rewards over sharing. (Medium confidence) \newline
(6) Loss Aversion \& Framing Effects – Framing the endowment in terms of losses rather than gains may influence lower generosity. (Medium confidence) \newline
(7) Cultural \& Demographic Variations – Conducting the experiment in a culture with lower altruism tendencies may reduce sharing but is harder to control. (Low confidence) \newline
(8) Incentive Structure Modification – Offering bonuses for minimizing sharing may reduce generosity but does not directly target underlying social or psychological behaviors. (Low confidence)
} \\
\midrule
4 & \makecell[{{p{15cm}}}]{
(1) Social Influence via Confederate – A confederate influences the dictator’s decision through nonverbal cues or suggestions. (Confidence: 0.9) \newline
(2) Incentivized Sharing – Offering a bonus for sharing a significant proportion may unexpectedly decrease sharing due to strategic decision-making. (Confidence: 0.9) \newline
(3) Variable Endowment Based on Performance – The dictator’s past economic performance determines their endowment, possibly altering fairness perceptions. (Confidence: 0.8) \newline
(4) Public Disclosure \& Social Repercussions – Informing the dictator that their decision will be publicly revealed could influence behavior. (Confidence: 0.6) \newline
(5) Narrative on Recipient’s Financial Struggles – Providing a story about the recipient’s economic conditions could impact generosity. (Confidence: 0.5)
} \\
\midrule
5 & \makecell[{{p{15cm}}}]{
(1) Variable Endowment – Adjusting the endowment size across rounds may influence perceptions of fairness and reduce sharing. (Confidence: 0.73) \newline
(2) Social Norms Influence – Playing multiple rounds with different partners may reinforce fairness norms but could also decrease sharing based on reciprocity. (Confidence: 0.67) \newline
(3) Feedback Mechanism – Providing immediate feedback on recipients’ reactions may influence decisions, potentially leading to lower generosity. (Confidence: 0.59) \newline
(4) Loss Aversion – Framing decisions as "saving" rather than "giving" could shift behavior by leveraging loss aversion tendencies. (Confidence: 0.56) \newline
(5) Emotional Manipulation – Emphasizing the impact of the dictator’s decision on the recipient’s well-being may alter behavior. (Confidence: 0.41) \newline
(6) Cognitive Load – Increasing decision complexity or adding multiple decision layers may lead to fatigue and reduced sharing. (Confidence: 0.38) \newline
(7) Cultural Context – Varying cultural backgrounds of participants could influence social norms of fairness and generosity. (Confidence: 0.34)
} \\
\bottomrule
\end{tabular}
\caption{Experiment designs outputted by \model 8B when asked to propose treatments that could decrease sharing behavior in the Dictator Game.}
\label{tab:results-designs-8b-dec}
\end{table}

\subsection{Inferring Knowledge}

\paragraph{Research workflow reasoning. }

The evaluation of this task follows the data formatting described in Appendix \ref{app:data-literature}. Specifically, we use 300 AER publications held out from training for evaluation. Two subtasks are considered: idea generation and title prediction. Following the evaluation metrics used in \citet{zhang2024massw}, we report BLEURT and ROUGE-1 scores. Full results, including those from expert models fine-tuned for this task, are presented in Table \ref{tab:results-sci-flow-full}.

\begin{table}[htbp]
\centering
\begin{tabular}{lcccc}
\toprule
 & \multicolumn{2}{c}{\textbf{BLEURT}($\uparrow$)} & \multicolumn{2}{c}{\textbf{ROUGE-1}($\uparrow$)} \\
\textbf{Model} & \textbf{Idea} & \textbf{Title} & \textbf{Idea} & \textbf{Title} \\
\midrule
\model 8B	&	0.498	&	0.438	&	0.076	&	0.406	\\
\model 70B	&	0.423	&	0.427	&	0.315	&	0.385	\\
\midrule
Llama 3.1 8B	&	0.460	&	0.441	&	0.083	&	0.076	\\
Llama 3.1 70B	&	0.482	&	0.398	&	0.106	&	0.238	\\
GPT 4o-mini	&	0.426	&	0.415	&	0.084	&	0.394	\\
GPT 4o	&	0.452	&	0.412	&	0.136	&	0.390	\\
\midrule
\model 8B$^*$	&	0.552	&	0.438	&	0.072	&	0.403	\\
\model 70B$^*$	&	0.379	&	0.409	&	0.254	&	0.362	\\
\bottomrule
\end{tabular}
\vspace{5pt}
\caption{
Performances on research workflow prediction. 
Given research contexts, the model-generated ideas are compared with the key ideas extracted from publication abstracts, while the generated titles are directly compared with ground truth paper titles. 
For each prediction scenario, BLEURT and ROUGE-1 scores are reported. Higher scores indicate better performances. 
``$^*$'' indicates \model models fine-tuned only on this task. 
}
\label{tab:results-sci-flow-full}
\end{table}

\paragraph{Complex problem solving. } 

To further evaluate \model’s capabilities in solving complex problems in behavioral science and economics, we use standard problems from the International Economics Olympiad (IEO)\footnote{Official website of the International Economics Olympiad (IEO): \url{https://ieo-official.org/}, retrieved April 9, 2025.}. 
These problems test the ability to represent, reason about, and apply domain-specific knowledge. 
We curate multiple-choice questions from five years of IEO contests (2019, 2021–2024), yielding a total of 91 questions. To ensure consistency, we include only questions that do not require reference to images.
For evaluation, each question is run 10 times in separate chat sessions. 

For each problem, we prompt the model with the following instruction:
\begin{framed}
\textbf{System prompt:}\\
    You are an expert in behavioral economics. Given a multiple-choice economics contest question, your task is to answer this question.
    
\textbf{User prompt:}\\
    \#\# Question:\\
    \qquad Topic: (topic)\\
    \qquad Question: (question)\\ \\
    \#\# Potential Choices:\\
    \qquad    A. (choice A)\\
    \qquad    B. (choice B)\\
    \qquad    C. (choice C)\\
    \qquad    D. (choice D) \\
    \qquad    Choose the correct answer from A, B, C, or D. Respond only with the letter. 
\end{framed}

The results are presented and discussed in Sec. \ref{sec:infer-knowledge} (Table \ref{tab:results-qa}).

\end{document}